\documentclass{article}

     \usepackage[preprint, nonatbib]{neurips_2020}

\usepackage[utf8]{inputenc} % allow utf-8 input
\usepackage[T1]{fontenc}    % use 8-bit T1 fonts
\usepackage{url}            % simple URL typesetting
\usepackage{booktabs}       % professional-quality tables
\usepackage{amsfonts}       % blackboard math symbols
\usepackage{nicefrac}       % compact symbols for 1/2, etc.
\usepackage{microtype}      % microtypography
\usepackage{xcolor}         % for text color
\usepackage{amsmath}
\usepackage{graphicx}
\usepackage{subfig}
\usepackage{float}
\usepackage{tcolorbox}
\usepackage[style=numeric, sorting=none]{biblatex}
\usepackage{amsthm}

\title{Geometric algorithms for predicting resilience and recovering damage in neural networks}

\author{
  Guruprasad Raghavan\\
  Caltech\\
  Pasadena, CA 91106 \\
  \texttt{graghava@caltech.edu} \\
  \And
  Jiayi Li\\
  UCLA \\
  Los Angeles, CA 90095 \\ 
  \texttt{jiayi.li@g.ucla.edu}
  \And
  Matt Thomson\\
  Caltech\\
  Pasadena, CA 91106 \\
  \texttt{mthomson@caltech.edu} \\
}

\begin{document}

\maketitle

\begin{abstract}
Biological neural networks have evolved to maintain performance despite significant circuit damage. To survive damage, biological network architectures have both intrinsic resilience to component loss and also activate recovery programs that adjust network weights through plasticity to stabilize performance. Despite the importance of resilience in technology applications, the resilience of artificial neural networks is poorly understood, and autonomous recovery algorithms have yet to be developed. In this paper, we establish a mathematical framework to analyze the resilience of artificial neural networks through the lens of differential geometry.  Our geometric language provides natural algorithms that identify local vulnerabilities in trained networks as well as recovery algorithms that dynamically adjust networks to compensate for damage. We reveal striking weight perturbation vulnerabilities in common  image analysis architectures, including MLP's and CNN's trained on MNIST and CIFAR-10 respectively. We also uncover high-performance recovery paths that enable the same networks to dynamically re-adjust their parameters to compensate for damage.  Broadly, our work provides procedures that endow artificial systems with resilience and rapid-recovery routines to enable their deployment for critical applications. 
%(176)
%function under difficult environmental conditions and can 

\end{abstract}
%While the computational capabilities of biological neural networks are often emphasized in artificial intelligence
%Brains self-assemble during organismal development, and have the ability to grow, morph, and auto-configure \cite{ackerman1992discovering, singer1986brain}.
\section{Introduction}
Brains are remarkable machines whose computational capabilities have inspired many breakthroughs in machine learning \cite{iliadis2020brain, fukushima1980neocognitron, wolczyk2019biologically, zador2019critique}. However, the resilience of the brain, its ability to maintain computational capabilities in harsh conditions and following circuit damage, remains poorly developed in current artificial intelligence paradigms \cite{hendrycks2019benchmarking} . 
Biological neural networks are known to implement redundancy and other architectural features that allow circuits to maintain performance following loss of neurons or lesion to sub-circuits
 \cite{gonzalez2019persistence, castor2018resilience, marder2006variability, srinivasan2011robustness, richter2019resilience}.  In addition to architectural resilience,  biological neural networks execute recovery programs that allow circuits to repair themselves through the activation of network plasticity following damage  \cite{arlotta2014brains, gates2000reconstruction,murphy2009plasticity}. For example, recovery algorithms reestablish olfactory and visual behaviors in mammals following sensory specific cortical circuit lesions \cite{keck2013synaptic,hong2018sensation}.  Through resilience and recovery mechanisms, biological neural networks can maintain steady performance in the face of dynamic challenges like changing external environments, cell damage, partial circuit loss as well as catastrophic injuries like the loss of large sections of the cortex. \cite{lewin1980your,scoville1957loss, hammerschmidt2015mice,gilbert2012adult}. 
 
Like brains, artificial neural networks must increasingly execute critical applications that require robustness to both hardware component damage and memory errors that could corrupt network weights. Recent studies have highlighted the importance of network robustness to soft errors that can lead to weight corruption and network failure \cite{li2017understanding} in applications including (i) decision-making in the healthcare industry, (ii) image and sensor analysis in self-driving cars and (iii) robotic control systems. Errors in dynamic access memory  can occur due to malicious attacks (the RowHammer), but a particular focus has been on errors induced by high energy particles \cite{arechiga2018robustness} that occur at surprising rates \cite{schroeder2009dram}.  Further, the rising implementation of neural networks on physical hardware (like neuromorphic, edge devices) \cite{monroe2014neuromorphic, mach2017mobile}, where networks can be disconnected from the internet and are under control of an end user, necessitates the need for damage-resilient and dynamically recovering artificial neural networks. 

The resilience of living neural networks motivates theoretical and practical efforts to understand the resilience of artificial neural networks and to design new algorithms that reverse engineer resilience and recovery into artificial systems \cite{firesmith_2019}.  Recent studies \cite{morcos2018importance, cheney2017robustness} have demonstrated empirically that MLP and CNN architectures can be surprisingly robust to large scale node deletion. However,  there is currently little understanding of the empirically observed resilience or what ultimately causes networks to fail. Mathematical frameworks will be important for understanding the resilience neural networks and for developing recovery procedures that can maintain network performance during damage. 

We propose a mathematical framework grounded in differential geometry for studying the resilience and the recovery of artificial neural nets. We formalize damage/response behavior as dynamic movement on a curved pseudo-Riemannian manifold. Our geometric language provides new procedures for identifying network vulnerabilities by predicting local perturbations that adversely impact the functional performance of the network. Further,  we demonstrate that geodesics, minimum length paths, on the weight manifold provide high performance recovery paths that the network can traverse to maintain performance while damaged.  Our algorithms allow networks to maintain high-performance during rounds of damage and repair through computationally efficient weight-update algorithms that do not require conventional retraining. Broadly, our work provides procedures that will help endow artificial systems with resilience and autonomous recovery policies to emulate the properties of biological neural networks.
\section{Analyzing network resilience with differential geometry}
We develop a geometric framework for understanding how artificial neural networks respond to damage using differential geometry to analyze changes in functional performance given changes in network weights. Two recent papers have highlighted intrinsic robustness properties of layered neural networks \cite{morcos2018importance, cheney2017robustness}. We provide a geometric approach for understanding robustness as arising from underlying geometric properties of the weight manifold that are quantified by the metric tensor. The geometric approach allows us to identify vulnerabilities in common neural network architectures as well as define new strategies for repairing damaged networks.

We represent a feed-forward neural network as a smooth,  $\mathbb{C}^\infty$function $f(\mathbf{x},\mathbf{w})$, that maps an input vector, $\mathbf{x} \in \mathbb{R}^{\text{k}}$, to an output vector, $f(\mathbf{x},\mathbf{w})  = \mathbf{y}\in \mathbb{R}^{\text{m}}$. The function, $f(\mathbf{x},\mathbf{w})$, is parameterized by a vector of weights, $\mathbf{w} \in \mathbb{R}^\text{n}$, that are typically set in training to solve a specific task. We refer to $W= \mathbb{R}^n$ as the \textit{weight space} ($W$) of the network, and we refer to $\mathcal{F} = \mathbb{R}^m$  as the \textit{functional manifold} \cite{mache2006trends}. In addition to $f$, we will sometimes be interested in considering a loss function, $L: \mathbb{R}^\text{m} \times\mathbb{R} \rightarrow \mathbb{R}$, that provides a scalar measure of network performance for a given task (Figure 1). 

We ask how the performance of a  trained neural network, $\mathbf{w_t}$, will change when subjected to weight perturbation, shifting $\mathbf{w_{trained}} \rightarrow  \mathbf{w_{damaged}}$. We use differential geometry to develop a mathematical theory, rooted in a functional notion of distance, to analyze how arbitrary weight perturbations $\mathbf{w_t} \rightarrow \mathbf{w_d}$ impact functional performance of a network.  Specifically, we construct a local distance metric, $\mathbf{g}$, that can be applied at any point in $W$ to measure the functional impact of an arbitrary network perturbation. 

To construct a metric mathematically,  we fix the input, $\mathbf{x}$, into a network and ask how the output of the network, $f(\mathbf{w},\mathbf{x})$, moves on the functional manifold, $\mathcal{F}$, given an infinitesimal weight perturbation, $\mathbf{du}$, in $W$ where $\mathbf{w_d} = \mathbf{w_t} + \mathbf{du}$.  For an infinitesimal perturbation $\mathbf{du}$, 
\begin{equation}
f(\mathbf{x},\mathbf{w_t} +\mathbf{du}) \approx f(\mathbf{x},\mathbf{w_t}) +\mathbf{ J_{w_t}} \ \mathbf{du},
\end{equation}

where $\mathbf{ J_{w_t}}$ is the Jacobian of $f(\mathbf{x},\mathbf{w})$ for a fixed $\mathbf{x}$, $J_{i,j} = \frac{\partial f_i}{\partial w^j}$, evaluated at $\mathbf{w_t}$. We measure the change in functional performance given $\mathbf{du}$ as the mean squared error     
\begin{align}
\label{local metric}
d(\mathbf{w_t},\mathbf{w_d})= |f(\mathbf{x},\mathbf{w_t})-f(\mathbf{x},\mathbf{w_d})|^2 &= \mathbf{du}^T \ (\mathbf{J_{w_t}}^T \ \mathbf{J_{w_t}}) \ \mathbf{du}\\
 &= \mathbf{du}^T \ \mathbf{g_{w_t}} \ \mathbf{du},\\
%&= \langle du,du \rangle_{w_t}
\end{align}where $\mathbf{g_{w_t}} = \mathbf{J_{w_t}}^T \mathbf{J_{w_t}}$ is the metric tensor evaluated at the point $\mathbf{w_t} \in W$. The metric tensor $\mathbf{g}$ is an $n \times n$ symmetric matrix that defines an inner product and local distance metric, $\langle \mathbf{du}, \mathbf{du} \rangle_{\mathbf{w}}  = \mathbf{du^T} \ \mathbf{g_{w}} \ \mathbf{du}$, on the tangent space of the manifold, $T_w(W)$ at each $\mathbf{w} \in W$.  

\begin{figure}[t]  
    \centering
  \centerline{  \includegraphics[scale=0.8]{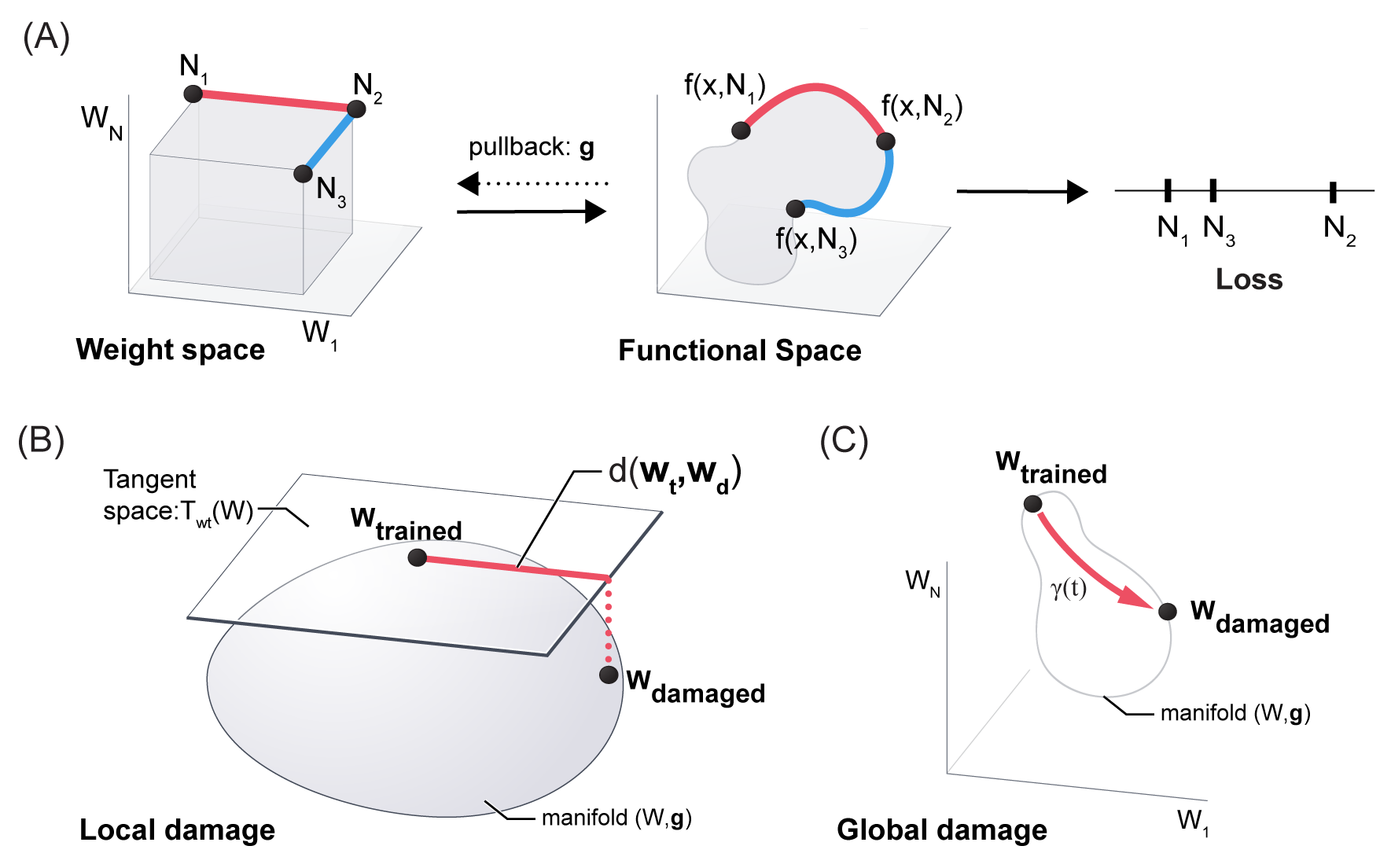}}
    \caption{\textbf{Geometric framework for analyzing neural network resilience} (A) Three networks ($N_1,N_2,N_3$) in weights space $W$ and their relative distance in functional space and loss space. Damage is analyzed by asking how movement in weight space changes functional performance and loss through introduction of a pullback metric $\mathbf{g}$. (B) We consider local damage to a network as an infinitesimal perturbation that can be analyzed in the tangent space of a trained network. (C) Global damage is modeled as long range movement of network weights along a path, $\gamma(t)$, in weight space. }
    \label{fig:schematics}
    \vspace{-2mm}
\end{figure}

Explicitly, 
\begin{equation}
    g_{ij} = \sum_{k=1}^{m}\frac{\partial f_k(\textbf{x},\mathbf{w})}{\partial \mathbf{w^i}}\frac{\partial f_k(\textbf{x},\mathbf{w})}{\partial \mathbf{w^j}},
\end{equation}where the partial derivatives $\frac{\partial f_k(\textbf{x},\mathbf{w})}{\partial \mathbf{w^i}}$ measure change in functional output of a network given a change in weight. In the appendix, we extend the metric formulation to cases where we  consider a set, $\mathbf{X}$, of training data and view $\mathbf{g}$ as the average of metrics derived from individual training examples. The metric, $\mathbf{g}$, provides a local measure of \textit{functional distance} on the pseudo-Riemmanian manifold $(W,\mathbf{g})$. At each point in weight space, the metric defines the length, $\langle \mathbf{du}, \mathbf{du} \rangle_{\mathbf{w}}$, of a local perturbation by its impact on the functional output of the network (Figure 1b).

Globally, we can use the metric to determine the functional performance change across a path connected set of networks. Mathematically, the metric changes as we move in $W$ due to the curvature of the ambient space that reflects changes in the vulnerability of a network to weight perturbation (Figure 1c).   As a network moves along a path, $\gamma(t) \in W$ from a given trained network $\gamma(0) = \mathbf{w_t}$ to a damaged network $\gamma(1) = \mathbf{w_d}$, we can analyze the integrated impact of damage on network performance along $\gamma(t)$ by using the metric to calculate the length of the path $\gamma(t)$ as: 
\begin{equation}
    L(\gamma) = \int^1 _0 {\langle \frac{d\gamma(t)}{dt}, \frac{d\gamma(t)}{dt}\rangle_{\gamma(t)}  }\ dt ,
    %L(\gamma) = \int_{t=0}^{1}{ \left[\frac{d\gamma(t)}{dt}\right]^T \mathbf{g}(\gamma(t)) \left[\frac{d\gamma(t)}{dt}\right]  dt } ,
    \label{eq:geodesicLength}
\end{equation}
where $\langle \frac{d\gamma(t)}{dt}, \frac{d\gamma(t)}{dt}\rangle_{\gamma(t)} = \frac{d\gamma(t)}{dt}^T \mathbf{g}_{\gamma(t)} \frac{d\gamma(t)}{dt}$ is the infinitesimal functional change accrued while traversing path $\gamma(t) \in W$. 

In what follows, we study the resilience of neural networks by analyzing the structure of the metric tensor along paths in weight space. We show that the metric tensor can be used to develop recovery procedures by finding `geodesic paths', minimum length paths, in the pseudo-Riemannian manifold that allow networks to respond to damage while suffering minimal performance degradation. 
\section{The geometry of local damage and network vulnerability}
We, first, apply our mathematical framework to analyze the response of trained neural networks to small, local weight perturbations. Empirical studies have demonstrated that trained networks are often robust to small, local weight perturbation \cite{morcos2018importance, cheney2017robustness}. We connect local resilience to the spectral properties of the metric tensor, $\mathbf{g}$, at a given position, $\mathbf{w_t}$, in weight space. We find that networks are typically robust to random local weight perturbations but also have catastrophic vulnerabilities to specific low magnitude weight perturbations that dramatically alter network performance. 

To understand local damage, we consider a trained network, $\mathbf{w_t}$, and we subject the network to an  infinitesimal weight perturbations in a direction $\mathbf{du} = c_i \ \mathbf{dw^i}$ yielding the perturbed weights $\mathbf{w'} = \mathbf{w_t} +\mathbf{du}$. We use $\mathbf{dw^i}$ to indicate an infinitesimal displacement vector in the direction $w_i$. Formally, we view $\mathbf{du}$ as a vector in the tangent space of W at $\mathbf{w_t}$, $T_{w_t}(W)$ (Figure 1B). The metric tensor, evaluated at the point $\mathbf{w_t}$ provides a local measure of functional performance change induced by the perturbation along $\mathbf{du}$  through Equation \ref{eq:func_eig}. 

\begin{figure}[t]
    \centering
    \includegraphics[scale=0.35]{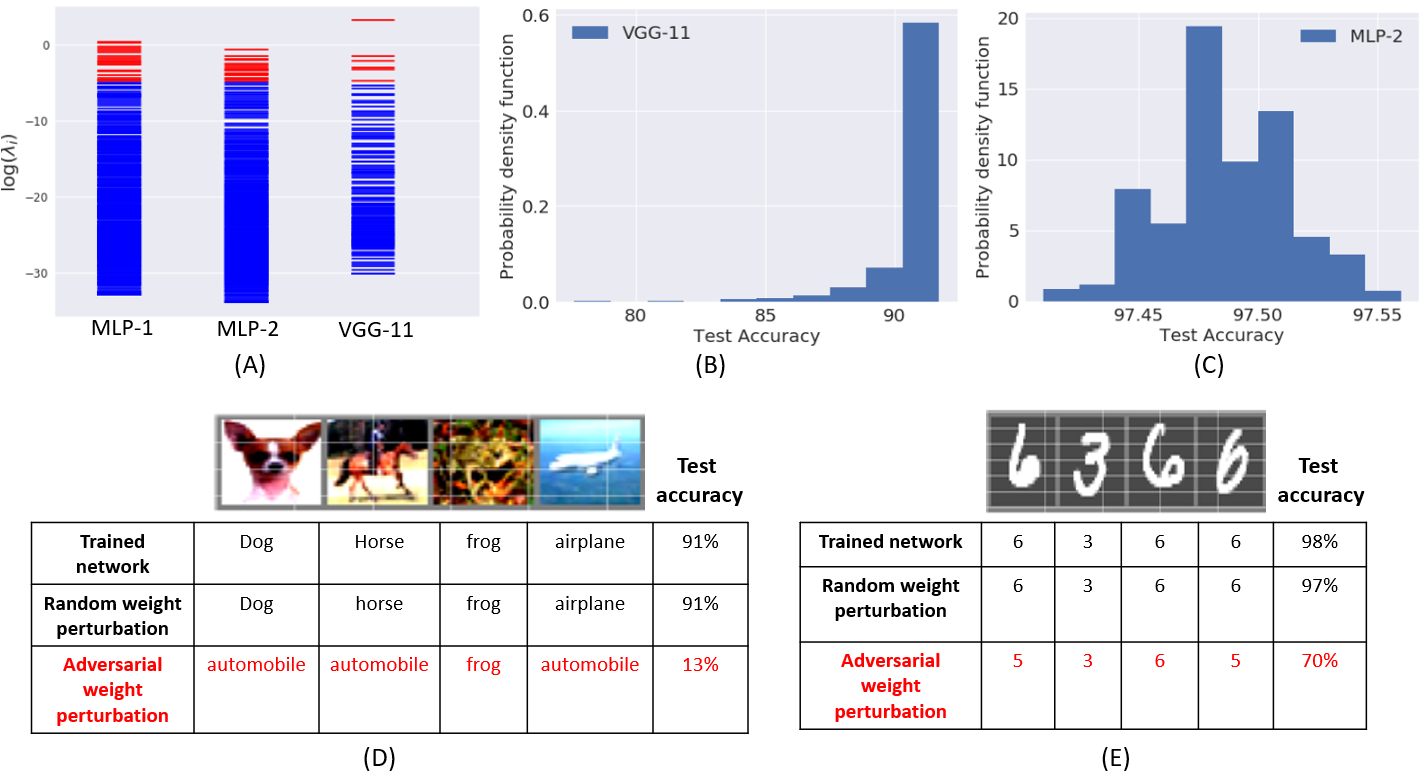}
    \caption{\textbf{Metric tensor explains local resilience and predicts catastrophic vulnerabilities.} (A) Spectra of the metric tensor for MLP-1, MLP-2 and VGG-11 (B,C) Test performance of networks perturbed within a unit-ball in $W$(B) perturbed VGG-11 trained on CIFAR-10, (C) perturbed MLP-2 trained on MNIST (D,E) Designing adversarial perturbations to destroy trained networks' performance. (D) adversarial perturbation within unit-ball in $W$ lowers accuracy to $13 \%$ in VGG-11, (E) adversarial weight perturbation within unit-ball in $W$ lowers accuracy to $70\%$ in MLP-2.}
    \label{fig:localDmg}
    \vspace{-6mm}
\end{figure}
As a positive semi-definite, symmetric matrix, $\mathbf{g}$  (evaluated at $\mathbf{w_t}$) has an orthonormal eigenbasis $\{ \mathbf{v_i} \}$ with eigenvalues $\lambda_i$, $\lambda_i \geq 0$.  The eigenvalue $\lambda_i$ locally determines how a perturbation along the eigenvector $\mathbf{v_i}$ will alter functional performance. Expanding an arbitrary perturbation, $\mathbf{du}$ in the basis $\{\mathbf{v_i}\}$, as $\mathbf{du} = \sum_i c_i \ \mathbf{v_i}$, the functional performance change of the network is
\begin{align}
d(\mathbf{w_t},\mathbf{w_t}+\mathbf{du}) & = \mathbf{du^T} \ \mathbf{g_{w_t}} \ \mathbf{du} \\
&= \sum_{i} c^2_i \lambda_i
\label{eq:func_eig}
\end{align}
where $c_i = \langle \mathbf{du} , \mathbf{v_i} \rangle$ quantifies the contribution of vector $\mathbf{v_i}$ to $\mathbf{du}$. Thus, the performance change, $d(\mathbf{w_t},\mathbf{w_t}+\mathbf{du})$,  incurred by a network, following perturbation $\mathbf{du}$ is determined by the magnitude of each $\lambda_i$ and the projection of $\mathbf{du}$ onto $\mathbf{v_i}$. The eigenvalues $\lambda$ convert weight changes into change in functional performance and so have units of $\frac{\text{performance change}}{\text{weight change}}$.  A network will be resilient to weight perturbations directed along eigenvectors, $\mathbf{v_i}$, with \textit{small} eigenvalues ($\lambda_i < 10^{-3}$). Alternately, networks are vulnerable to perturbations along directions with larger eigenvalues ($\lambda_i > 10^{-3}$).  Our definition of resilient directions, $\lambda_i < 10^{-3}$, is an operational direction that selects directions where a unit of weight change will produce a performance change of less than $10^{-3}$ or $.1\%$. 

Mathematically, we can understand the resilience of networks to randomly distributed weight perturbations by calculating the average response of a network to Gaussian weight perturbations, $\mathbf{du} \sim P(\mathbf{du}) $, where $P(du_i) =\mathcal{N}(0,\tfrac{\sigma}{d})$  ($n = \text{dim}(W)$  and $\mathbb{E}[||\mathbf{du}||_2 ]  \  = \sigma)$ .  The expectation of the induced performance change for such a Gaussian perturbation is
\begin{align}
\mathbb{E}_{du_i \sim \mathcal{N}(0,\tfrac{\sigma}{d})  } [d(\mathbf{w},\mathbf{w}+\mathbf{du})  ]&=  \frac{\sigma}{d} \sum_i \lambda_i \\
&< \ \sigma \ \rho \ \lambda_1, 
\end{align}where $\rho$ indicates the fraction of vulnerable directions, and $\lambda_1$ is the largest eigenvalue of $\mathbf{g}$.  

Empirically, we find that trained networks are, perhaps as expected, robust to `random' local perturbation (Figure 2) due to a large fraction of resilient eigendirections ($\rho < 10^{-3}$). Such local network robustness holds for a series of trained network architectures including (i) Multi-layer perceptrons (MLP-1, MLP-2) trained on MNIST and (ii) Convolutional neural networks (VGG-11) trained on CIFAR-10. MLP-1 is a single hidden layer network, with variable number of hidden nodes, while MLP-2 is the LeNet architecture borrowed from \cite{lecun1998gradient} (2 hidden layers, with 300 and 100 hidden nodes respectively). VGG-11 for CIFAR-10 is adapted from \cite{simonyan2014very}\footnote{The network architecture, pre-trained models and optimization algo's are specified in the appendix}.

Consistent with their eigenspectra (VGG-11: $ \rho < 10^{-4}$, MLP-1, MLP-2: $\rho< 10^{-3}$) ,  both MLP and CNN architectures exhibit minimal performance degradation for unit-ball perturbations \footnote{Unit-ball perturbations, due to the high dimensionality of the space, induce an average weight change of $<10^{-6}$ for individual weights } ($\sigma = 1$, Figure 2A) .  When perturbed along 1000 directions of unit-norm, the trained MLP-2 (initial test accuracy of 98$\%$) maintains accuracy of 97.2-97.6$\%$ (Figure 2C). Perturbation of VGG-11 trained on CIFAR-10 (initial test accuracy of 91$\%$) yields  networks with test accuracy between 88-91$\%$ (Figure 2B). 

Resilience to such small local perturbations might be expected, but our framework also exposes hidden catastrophic vulnerabilities to perturbations of the same order in both networks.  By designing adversarial weight perturbations to lie along the `vulnerable' eigenvectors of $\mathbf{g}$ ($\mathbf{v_i}$ with large $\lambda_i$),  we can induce sharp performance declines across architectures (Figure 2D,E). For the VGG-11 network trained on CIFAR-10, an adversarial weight perturbation decreases accuracy from 91$\%$ to 13$\%$ (Figure 2D). Similarly, adversarial perturbation reduces the performance of MLP-2 network trained on MNIST from 98$\%$ to 70$\%$ (Figure 2E). For the CIFAR-10 network, a relatively small perturbation causes the network to make critical classification errors making the erroneous inference of most CIFAR-10 images to being in the class of `automobiles'. In this way, the local geometry of the weight manifold allows us to discover subtle weight perturbations that cause catastrophic changes in network performance for small change in network weights. 

\section{Acceleration identifies global break-down points in a network}
From earlier studies \cite{morcos2018importance, cheney2017robustness}, we know that trained MLP's and CNN's can be surprisingly robust to much more profound global damage including large scale node deletion.  In this section, we develop a concept of break-down acceleration using the \textit{covariant derivative} of a network along paths connecting the trained network and damaged network in $W$.  Break-down acceleration  \textit{predicts} failure points that emerge in weight space through rapid changes in the curvature of the weight space, and ultimately allows us to develop procedures to thwart break-down by avoiding acceleration. 

\begin{figure}[t]
    \vspace{-3mm}
    \centering
    \subfloat[]{{\includegraphics[scale=0.4]{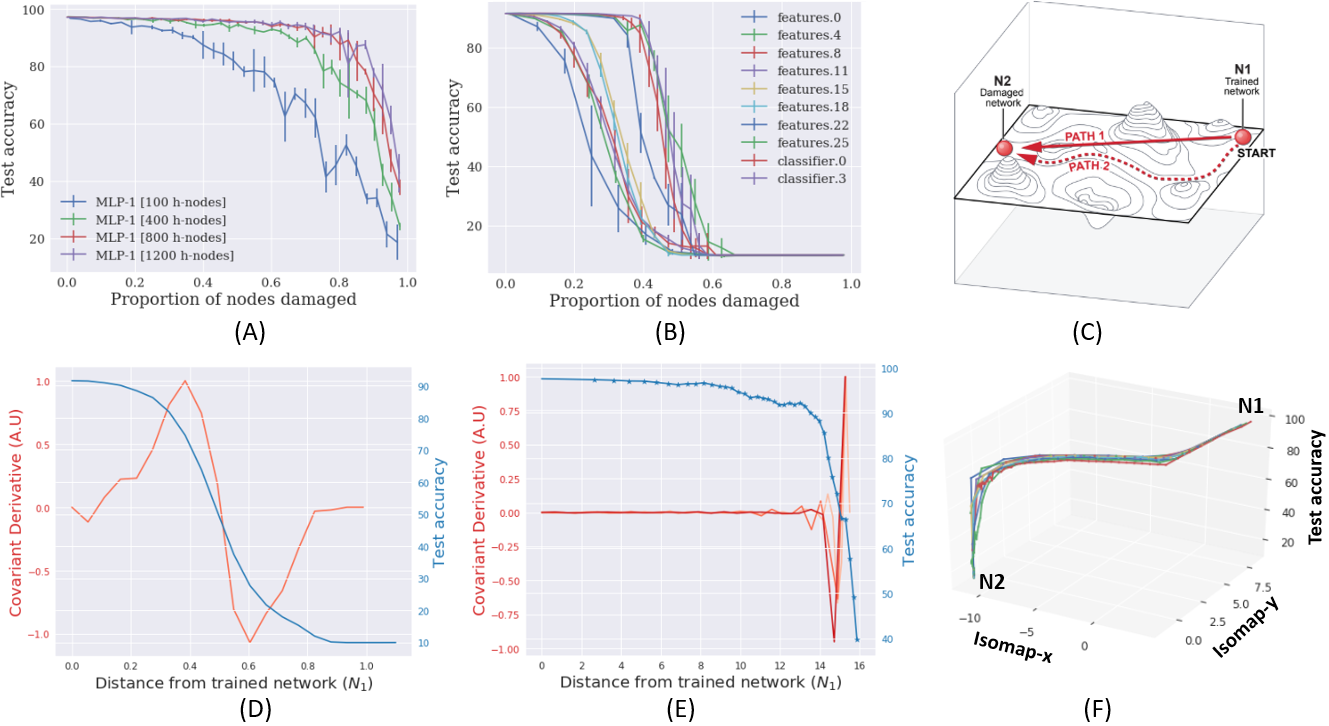}}}\hfill
    \vspace{-3mm}
    \caption{ \textbf{Break-down acceleration characterizes network break-down points following damage} Performance of an (A) MLP-1 network (1 hidden-layer, variable hidden nodes) and (B) VGG-11 during simulated damage to distinct layers. Both networks experience sharp performance break down when network damage exceeds (A) $\sim$90$\%$ of hidden-nodes for MLP-1 and (B) $\sim$60$\%$ of nodes in any layer for VGG-11. (C,D,E) \textbf{Damage paths in manifold ($W,g$).} (C) A cartoon of the loss landscape showing multiple break-down paths from the trained network to the damaged network (D,E) The covariant derivative of the accuracy along multiple damage paths for (D) MLP-2 and (E) VGG-11 are shown. A steep increase in the covariant derivative (acceleration) along damage paths corresponds to the networks' sharp break-down to global damage.(F) Multiple damage paths (colored lines) shown from trained MLP-2 (N1) to its damaged counter-part (N2). The z-axes is the test-accuracy of the networks, while x,y axes are the isomap embedding of networks in a 2D space. }
    \vspace{-3mm}
    \label{fig:globalDamage}
\end{figure}
Mathematically,  we represent global damage as a path in weight space, $\gamma{(t)} \in W$ with  $ t \in [0,1]$, that connects a trained network, $\gamma(0) = \mathbf{w_t}$, to its damaged counterpart $\gamma(1) = \mathbf{w_d}$ (Figure 3C). Practically, global damage might emerge as a discrete event (node deletion), our analysis provides a continuous approximation to discrete network damage. As a network moves along a path from $\mathbf{w_t}$ to $\mathbf{w_d}$, the metric tensor itself changes, changing its spectra and its vulnerability. 

Along a path, $\gamma(t) \in W$ the velocity vector,  $v(t) = \frac{d\gamma}{dt}$, quantifies the change in the functional performance of a network per unit time. Mathematically, we define the break-down speed ($s$) of a network along a path in weight space as the norm of the network's velocity vector computed using the metric tensor $s(t) = \langle \frac{d\gamma}{dt},  \frac{d\gamma}{dt} \rangle_{\gamma(t)} =\sum_{i j} g_{i j} w_{t i}\ w_{t j}$. Non-linear break-down points emerge along paths in $W$ when break-down speed undergoes a rapid acceleration, so that $\frac{ds}{dt} >> 0$.  We can calculate the break-down speed and acceleration explicitly for a network following simple straight or Euclidean path from a trained to damaged configuration. Taking $w_d =0 $, $\gamma(t) = \mathbf{w_t}(1-t) $  and $\frac{d \gamma(t)}{dt} = -\mathbf{w_t}$, we have
 \begin{align}
%&\langle \frac{d\gamma}{dt},  \frac{d\gamma}{dt} \rangle_{\gamma(t)} =\sum_{i j} g_{i j} w_{t i}\ w_{t j} \\
 &\frac{ds}{dt} = \sum_{i,j} \sum_k \frac{d g_{ij}}{dw_k}  w_{t k} \ w_{t i}\ w_{t j}
 \end{align}where $g_{i j}$ is evaluated along $\gamma(t)$. The change in the metric tensor $\frac{d g_{ij}}{dw_k}$ along a path $\gamma(t)$, thus, determines whether performance decays at a constant $\frac{ds}{dt} = 0$, $\frac{dg_{i j}}{dk}=0$ or at an accelerating $\frac{ds}{dt} >0$, $\frac{dg_{i j}}{dw_k}>0$ rate.  For curved paths break-down acceleration can be analyzed using an object known as the covariant derivative, $\mathbf{\nabla}_{\gamma(t)} v(t)$ (Appendix). 

In practice, calculation of the break-down acceleration identifies, damage failure points in real neural networks.  For example, both MLP-1 and VGG-11 architectures tolerate considerable node deletion (Figure 3A). MLP-1, (one hidden layer, 400 hidden units) trained on MNIST, tolerates damage to 80$\%$ of the network nodes reducing  functional performance of the network by merely $\sim$10$\%$.  Similarly, VGG-11 trained on CIFAR-10 tolerates $\sim 60\%$ node damage in any layer without performance degradation. However, both networks exhibit drastic break-down in functional performance beyond these node damage thresholds (Figure 3A,B).  Mathematically, break-down points occur where the acceleration of the network, as measured by the covariant derivative along the damage path, rapidly increases  (Figure 3D-F)  Steep increases in the covariant derivative identify points of loss acceleration corresponding to the functional breakdown of both the networks analyzed. 

\section{Geodesic paths enable network recovery}
Thus,  globally, network break-down occurs along a damage path in $W$due to abrupt changes in the curvature of the underlying functional landscape that result in abrupt change in the metric. The mathematical connection between break-down and curvature suggests a strategy for designing recovery protocols that can adapt a neural network's weights to compensate for damage. Inspired by recovery mechanisms in neuroscience that compensate for damage by altering the weights of undamaged nodes. We apply the concept of break-down acceleration to develop recovery procedures for artificial neural networks that compensate for damage through continuous adjustment of the undamaged weights by minimizing the acceleration along the path. 

Mathematically, minimum acceleration paths in weight space are known as geodesic paths. Geodesic paths, by definition,  provide both minimum length and minimum acceleration paths in weight space. Specifically, we consider a trained network, $\mathbf{w}$, subjected to weight damage that zeros a subset of weights, $w_i=0 $, for $i \in n_{\text{damaged}}$.  Our strategy responds to damage by adjusting undamaged weights, $w_i$ for $i \notin n_{\text{damaged}}$ to maximize network performance by moving the network along a geodesic in $W$. Geodesic paths can be computed directly using our metric $\mathbf{g}$ and also represent the minimum distance paths (with distance defined in equation-\ref{eq:geodesicLength}) between two points on $W$. 
%To find geodesic recovery paths on $W$, we can solve the geodesic equation 
%\begin{align}
   % \frac{d^2 w^\eta}{dt^2} + \Gamma_{\mu\nu}^\eta\frac{dw^\mu}{dt} \frac{dw^\nu}{dt} = 0
    %\label{eq:geodesicMotion}
%\end{align}
%where, $w^j$ defines the j'th basis vector of the weights space $W$, $\Gamma_{\mu\nu}^\eta$ specifies the Christoffel symbols $(  \Gamma_{\mu \nu}^\eta =  \sum_r \frac{1}{2}g_{\eta r}^{-1}(\frac{\partial g_{r \mu}}{\partial x^\nu} + \frac{\partial g_{r \nu}}{\partial x^\mu} - \frac{\partial g_{\mu \nu}}{\partial x^r}))$ on the manifold. The Christoffel symbols capture infinitesimal changes in the metric tensor ($\mathbf{g}$) along a set of directions in the manifold. They are computed by setting the covariant derivative of the metric tensor along a path specified by $\gamma(t)$ to zero.  We, specifically, compute geodesic paths, $\gamma(t)$, so that $\gamma(0) = \mathbf{w_t}$ and $\gamma(1) \in W_d$ where $W_d$ is the damage hyper-plane. The damage hyper-plane is the set,  $W_d= \{w_i = 0, \ \forall i \ \in n_\text{damage} \} \subset W$, of all networks that are consistent with a given configuration of weight damage. Thus, we find paths through weight space that achieve a given configuration of damage while maximizing network performance. 
\begin{figure}[t]
    \centering
    \subfloat[\label{fig:recoveryNetwork}]{{\includegraphics[scale=0.45]{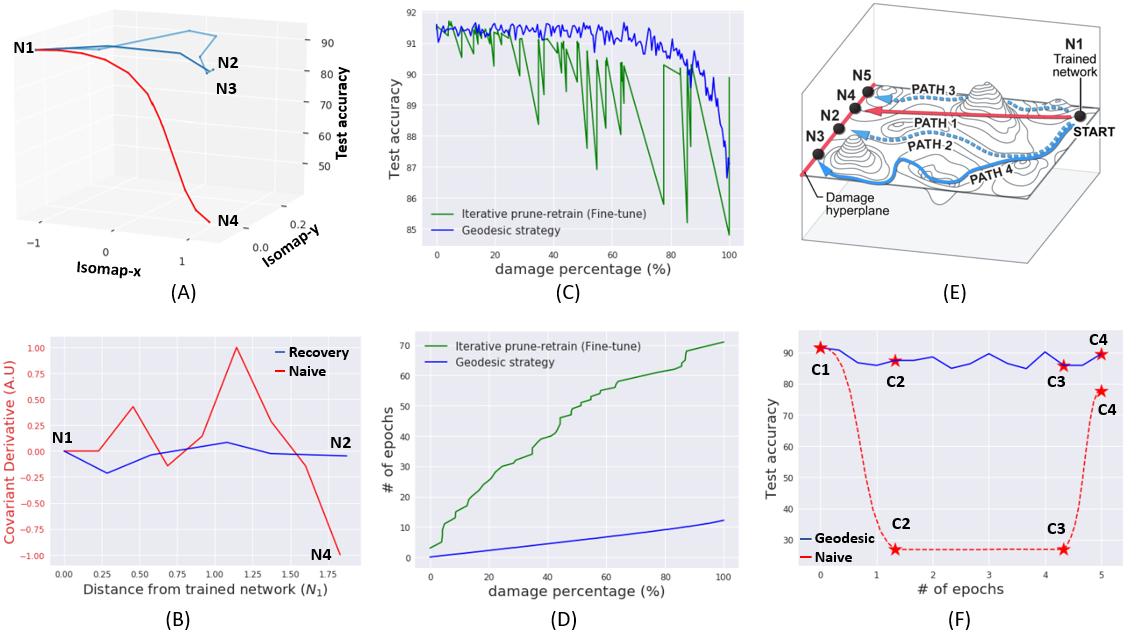}}}\hfill
    \vspace{-3mm}
    \caption{\textbf{Geodesic paths allow damage compensation through weight adjustment}: \textbf{(A)} Test accuracy of geodesic recovery paths (blue) versus naive damage paths (red) for VGG-1 network while $30$ convolution filters and 1000 nodes from fully-connected layers are damaged. While naive path exhibits sharp break-down, geodesic procedure adjusts undamaged weights to maintain test accuracy. \textbf{(B)} Magnitude of the covariant derivative (break-down acceleration) for geodesic (blue) and naive damage paths (red). \textbf{C} Test accuracy (C) and \textbf{(D)} number of network update epochs (D) for geodesic recovery (blue) vs fine-tuning (green) while  50 (out of 60) conv-filters are deleted from layer1 in VGG11. Geodesic recovery requires $\leq$10 total update epochs .\textbf{(E)} A depiction of multiple recovery paths on the loss landscape from trained network (N1) to networks on the damage hyper-plane (N2, N3, N4, N5). The z-axes is network-loss, while the x,y axes are neural net weights. \textbf{(F)} Geodesic strategy (blue) allows networks to dynamically transition between configurations: C1, trained VGG11 network ; C2,  50 conv-filters removed from C1; C3, 1000 additional nodes removed from classifier-layers in C2; C4, 30 conv-filters in conv-layer1 restored to C3.  Dynamic transitioning enabled within 5 epochs. Naive strategy is red).}
    \label{fig:rapidRecovery}
    \vspace{-3mm}
\end{figure}
In general, geodesic paths are typically calculated using the geodesic equation (Appendix) an ordinary differential equation that uses derivatives of the metric tensor to identify minimum acceleration paths in a space given an initial velocity. However, solutions to the geodesic equation are computationally prohibitive for large neural networks as they require evaluation of the Christoffel symbols which scale as a third order polynomial of the number of parameters in the neural network ($\mathcal{O}(n^3)$). 

Therefore, we developed an approximation to the geodesic equation using a first order expansion of the loss function.  Given a trained network, our procedure updates the weights of the network to optimize performance given a direction of damage. To discover a geodesic path $\gamma(t)$, we begin at a trained network and iteratively solve for the tangent vector , $\theta(w)$, at every point, $\mathbf{w} = \gamma(t) $, along the path, starting from $\mathbf{w_t}$ and terminating at the damage hyperplane, $W_d$. The damage hyperplane  is the set of all networks, $\mathbf{w} \in W$, such that $w_i=0 $, for $i \in n_{\text{damaged}}$. We specifically solve
\begin{equation}
    \text{argmin}_{\theta(\mathbf{w})}  \ \langle \theta(\mathbf{w}), \theta(\mathbf{w}) \rangle_\mathbf{w} - \beta \ \theta(\mathbf{w})^T v_w \\ \text{      subject to:    } \theta(\mathbf{w})^T \theta(\mathbf{w}) \leq 0.01.
\end{equation}The tangent vector $\theta(\mathbf{w})$ is obtained by simultaneously optimizing two objective functions: (1) minimizing the increase in functional distance along the path measured by the metric tensor ($\mathbf{g_w}$) [min: ($\langle \theta(\mathbf{w}), \theta(\mathbf{w}) \rangle_\mathbf{w}$) = $ (\theta(\mathbf{w})^T\mathbf{g_w}\theta(\mathbf{w}))$ ] and (2) maximizing the dot-product between the tangent vector and $v_\mathbf{w}$, vector pointing in the direction of the hyperplane [max: ($\theta(\mathbf{w})^T v_\mathbf{w}$)] to enable movement towards the damage hyperplane.    By finding geodesic paths to the damage hyperplane, we find weight adjustments that can be made within a network during damage to maintain performance (Figure 4E). 

Our optimization procedure is a quadratic program that trades off, through the hyper-parameter $\beta$,  motion towards the damage hyper-plane and the maximization of the functional performance of the intermediate networks along the path (optimization procedure elaborated in the appendix). The strategy discovers multiple paths from the trained network $\mathbf{w_t}$ to $W_d$, damage hyper-plane, (depicted as path-1 to path-5 in Figure 4a) where networks maintain high functional performance during damage. Of the many paths obtained, we can select the path with the shortest total length (with respect to the metric $\mathbf{g}$) as the best approximation to the geodesic in the manifold. 

The geodesic strategy enables damage compensation through continuously updating weights in the network. We apply the geodesic strategy to discover recovery paths from a trained network (VGG11) to a pre-defined damage hyperplane\footnote{Fig-4A, 4B: Damage hyperplane for VGG11 is defined by := deletion of 30 conv-filters from layer1,2 and 1000 nodes from fully-connected layer1,2.}. The recovery path is a high-performance path with all networks performing above 87$\%$ test accuracy, and the recovery path maintains low break-down acceleration when compared to the naive (linear) path (Figure 4B). A similar analysis for MLP's is presented in the appendix. 

While high-performance paths can also be discovered through heuristic fine-tuning, the geodesic procedure is both rationale and computationally efficient. Specifically, an iterative prune-train cycle achieved through structured pruning of a single node at a time, coupled with SGD re-training \cite{han2015learning, frankle2018lottery} (Figure 4C) requires 70 training epochs to identify a recovery path. In comparison,  the geodesic strategy finds paths that quantitatively out-perform the iterative prune-train procedure and obtains these paths with only 10 training epochs (figure 4C,D). 

Additionally,  the same geodesic strategy enables us to dynamically shift networks between different weight configurations (eg from a dense to sparse or vice-versa) while maintaining performance (Figure 4F). The rapid shifting of networks is relevant for networks on neuromorphic hardware to ensure that the real-time functionality of the hardware isn't compromised while transitioning between different power configurations. 
\section{Discussion}
We have established a mathematical framework to analyze resilience of neural networks through the lens of differential geometry. We introduce a functional distance metric on a Riemmanian weight manifold and apply the metric tensor, covariant derivative, and the geodesic to predict the response of networks to local and global damage. Mathematically, our work forms new connections between machine learning and differential geometry. 
Practically, we develop new procedures for (i) identifying vulnerabilities in neural networks and (ii) compensating for network damage in real-time through computationally efficient weight updates, enabling their rapid recovery. As neural networks are increasingly deployed on edge devices with increased susceptibility to damage, we believe these methods could be useful in a variety of practical applications.

%Our work also provides insight into resilience of biological networks. The observation that juvenile song-birds can tolerate large neuronal losses while adults cannot can be reconciled with our recovery paradigm. Our paradigm finds high-performance paths while networks are getting damaged due to constant synaptic plasticity. It is well-known that juvenile song-birds have high levels of plasticity and as a result can tolerate damage without facing deterioration in functional performance, while adults have limited plasticity.

\section{Broader Impact}

The field of AI has grown by leaps and bounds in the last few years. As a result, AI is increasingly being built into many critical applications across the society. Additionally, to cater to the rising need of AI systems for real-time applications, AI systems have been transitioning from cloud-implementation to edge devices and neuromorphic hardware. Some of the real-time critical applications that have actively adopted AI systems include (1) decision making in the health-care industry, (2) real-time image and sensor analysis in self-driving cars, (3) incorporation into IoT sensors and devices installed in most households and (4) robotic control systems. 

The failure of AI in any of these applications could be catastrophic. For instance, errors committed by AI systems while classifying radiology reports in the health-care industry, or the faulty real-time analysis of stream of images being processed by AI systems in self-driving cars could lead to human casualties. Hence, it has become extremely important for us to understand how neural network architectures (performing critical applications) react to perturbations, that could arise from many sources. AI implemented on the cloud are a victim of DRAM (dynamic random access memory) errors that can occur at surprising rates, either due to malicious attack or induced by high energy particles. Additionally, the growing implementation of AI networks on physical hardware (for instance, neuromorphic, edge devices) has made the need for discovering damage-resilient networks and rapidly recovery damaged networks a necessity.

Our paper lays down the mathematical framework to study resilience and robustness of neural networks to damage and proposes algorithms to rapidly recover networks experiencing damage. Our research will be extremely important for AI systems implemented across many applications, as damage of systems is inevitable and needs to be protected against. Although resilience and robustness of AI systems is very important, there aren't very many principled studies on the same. To reduce the gap in our knowledge on the resilience of AI, we propose a principled framework to understand the vulnerabilites of AI networks. One of the immediate applications of our contribution would be the design of damage-resilient networks and rapid recovery algorithms implemented on neuromorphic hardware. We believe this research will be foundational as neural networks are becoming ubiquitous across many applications, ranging from rovers sent to mars to radiology applications.

%focuses on studying robustness of neural networks through the lens of differential geometry and proposes algorithms to rapidly recover damaged networks as well as dynamically shift neural networks between different configurations.

%as AI becomes built into machine critical applications across society we need to understand how network architectures react to perturbation--from many sources. We also need to develop new classes of algorithms that can modulate networks to help them withstand component damage and loss. I feel like this will be very important in many many applications, but hasnt been considered that deeply in the literature. It been considered in biological evolution where resilience and recovery are critically important.

%I feel like this research will be foundational as neural networks become ubiquitous --self-driving cars, robots going to mars, radiology applications.

%Firstly, as neural networks are increasingly being implemented on physical hardware (like neuromorphics), rapid rescue of damaged networks becomes essential as this would enhance robustness of networks and lead to their quick integration with IoT devices. Secondly, neural networks have fallen victims to adversarial attacks, questioning their reliability and their usefulness for critical applications. As our recovery algorithm enables the discovery of paths consisting of high-performance networks  in the manifold 

%http://yann.lecun.com/exdb/publis/pdf/simard-00.pdf
%amari

\printbibliography

@inproceedings{li2017understanding,
  title={Understanding error propagation in deep learning neural network (DNN) accelerators and applications},
  author={Li, Guanpeng and Hari, Siva Kumar Sastry and Sullivan, Michael and Tsai, Timothy and Pattabiraman, Karthik and Emer, Joel and Keckler, Stephen W},
  booktitle={Proceedings of the International Conference for High Performance Computing, Networking, Storage and Analysis},
  pages={1--12},
  year={2017}
}

@inproceedings{arechiga2018robustness,
  title={The robustness of modern deep learning architectures against single event upset errors},
  author={Arechiga, Austin P and Michaels, Alan J},
  booktitle={2018 IEEE High Performance extreme Computing Conference (HPEC)},
  pages={1--6},
  year={2018},
  organization={IEEE}
}

@article{murphy2009plasticity,
  title={Plasticity during stroke recovery: from synapse to behaviour},
  author={Murphy, Timothy H and Corbett, Dale},
  journal={Nature Reviews Neuroscience},
  volume={10},
  number={12},
  pages={861--872},
  year={2009},
  publisher={Nature Publishing Group}
}

@article{srinivasan2011robustness,
  title={Robustness and fault tolerance make brains harder to study},
  author={Srinivasan, Shyam and Stevens, Charles F},
  journal={BMC biology},
  volume={9},
  number={1},
  pages={46},
  year={2011},
  publisher={Springer}
}

@article{hong2018sensation,
  title={Sensation, movement and learning in the absence of barrel cortex},
  author={Hong, Y Kate and Lacefield, Clay O and Rodgers, Chris C and Bruno, Randy M},
  journal={Nature},
  volume={561},
  number={7724},
  pages={542--546},
  year={2018},
  publisher={Nature Publishing Group}
}

@article{keck2013synaptic,
  title={Synaptic scaling and homeostatic plasticity in the mouse visual cortex in vivo},
  author={Keck, Tara and Keller, Georg B and Jacobsen, R Irene and Eysel, Ulf T and Bonhoeffer, Tobias and H{\"u}bener, Mark},
  journal={Neuron},
  volume={80},
  number={2},
  pages={327--334},
  year={2013},
  publisher={Elsevier}
}

@article{hendrycks2019benchmarking,
  title={Benchmarking neural network robustness to common corruptions and perturbations},
  author={Hendrycks, Dan and Dietterich, Thomas},
  journal={arXiv preprint arXiv:1903.12261},
  year={2019}
}

@inproceedings{han2015learning,
  title={Learning both weights and connections for efficient neural network},
  author={Han, Song and Pool, Jeff and Tran, John and Dally, William},
  booktitle={Advances in neural information processing systems},
  pages={1135--1143},
  year={2015}
}

@article{frankle2018lottery,
  title={The lottery ticket hypothesis: Finding sparse, trainable neural networks},
  author={Frankle, Jonathan and Carbin, Michael},
  journal={arXiv preprint arXiv:1803.03635},
  year={2018}
}

@article{gilbert2012adult,
  title={Adult visual cortical plasticity},
  author={Gilbert, Charles D and Li, Wu},
  journal={Neuron},
  volume={75},
  number={2},
  pages={250--264},
  year={2012},
  publisher={Elsevier}
}

@article{zador2019critique,
  title={A critique of pure learning and what artificial neural networks can learn from animal brains},
  author={Zador, Anthony M},
  journal={Nature communications},
  volume={10},
  number={1},
  pages={1--7},
  year={2019},
  publisher={Nature Publishing Group}
}

@article{fukushima1980neocognitron,
  title={Neocognitron: A self-organizing neural network model for a mechanism of pattern recognition unaffected by shift in position},
  author={Fukushima, Kunihiko},
  journal={Biological cybernetics},
  volume={36},
  number={4},
  pages={193--202},
  year={1980},
  publisher={Springer}
}

@article{wolczyk2019biologically,
  title={Biologically-Inspired Spatial Neural Networks},
  author={Wo{\l}czyk, Maciej and Tabor, Jacek and {\'S}mieja, Marek and Maszke, Szymon},
  journal={arXiv preprint arXiv:1910.02776},
  year={2019}
}

@article{marder2006variability,
  title={Variability, compensation and homeostasis in neuron and network function},
  author={Marder, Eve and Goaillard, Jean-Marc},
  journal={Nature Reviews Neuroscience},
  volume={7},
  number={7},
  pages={563--574},
  year={2006},
  publisher={Nature Publishing Group}
}

@article{cheney2017robustness,
  title={On the robustness of convolutional neural networks to internal architecture and weight perturbations},
  author={Cheney, Nicholas and Schrimpf, Martin and Kreiman, Gabriel},
  journal={arXiv preprint arXiv:1703.08245},
  year={2017}
}

@article{morcos2018importance,
  title={On the importance of single directions for generalization},
  author={Morcos, Ari S and Barrett, David GT and Rabinowitz, Neil C and Botvinick, Matthew},
  journal={arXiv preprint arXiv:1803.06959},
  year={2018}
}

@article{schroeder2009dram,
  title={DRAM errors in the wild: a large-scale field study},
  author={Schroeder, Bianca and Pinheiro, Eduardo and Weber, Wolf-Dietrich},
  journal={ACM SIGMETRICS Performance Evaluation Review},
  volume={37},
  number={1},
  pages={193--204},
  year={2009},
  publisher={ACM New York, NY, USA}
}

@article{richter2019resilience,
  title={Resilience to adversity is associated with increased activity and connectivity in the VTA and hippocampus},
  author={Richter, Anja and Kr{\"a}mer, Bernd and Diekhof, Esther K and Gruber, Oliver},
  journal={NeuroImage: Clinical},
  volume={23},
  pages={101920},
  year={2019},
  publisher={Elsevier}
}

@article{hammerschmidt2015mice,
  title={Mice lacking the cerebral cortex develop normal song: insights into the foundations of vocal learning},
  author={Hammerschmidt, Kurt and Whelan, Gabriela and Eichele, Gregor and Fischer, Julia},
  journal={Scientific reports},
  volume={5},
  pages={8808},
  year={2015},
  publisher={Nature Publishing Group}
}

@article{scoville1957loss,
  title={Loss of recent memory after bilateral hippocampal lesions},
  author={Scoville, William Beecher and Milner, Brenda},
  journal={Journal of neurology, neurosurgery, and psychiatry},
  volume={20},
  number={1},
  pages={11},
  year={1957},
  publisher={BMJ Publishing Group}
}

@article{lewin1980your,
  title={Is your brain really necessary?},
  author={Lewin, Roger},
  journal={Science},
  volume={210},
  number={4475},
  pages={1232--1234},
  year={1980},
  publisher={JSTOR}
}

@article{gonzalez2019persistence,
  title={Persistence of neuronal representations through time and damage in the hippocampus},
  author={Gonzalez, Walter G and Zhang, Hanwen and Harutyunyan, Anna and Lois, Carlos},
  journal={Science},
  volume={365},
  number={6455},
  pages={821--825},
  year={2019},
  publisher={American Association for the Advancement of Science}
}

@misc{iliadis2020brain,
  title={Brain-inspired computing and machine learning},
  author={Iliadis, Lazaros S and Kurkova, Vera and Hammer, Barbara},
  year={2020},
  publisher={Springer}
}

@book{mache2006trends,
  title={Trends and Applications in Constructive Approximation},
  author={Mache, Detlef H and Szabados, J{\'o}zsef and de Bruin, Marcel G},
  volume={151},
  year={2006},
  publisher={Springer Science \& Business Media}
}

@article{mach2017mobile,
  title={Mobile edge computing: A survey on architecture and computation offloading},
  author={Mach, Pavel and Becvar, Zdenek},
  journal={IEEE Communications Surveys \& Tutorials},
  volume={19},
  number={3},
  pages={1628--1656},
  year={2017},
  publisher={IEEE}
}

@misc{monroe2014neuromorphic,
  title={Neuromorphic computing gets ready for the (really) big time},
  author={Monroe, Don},
  year={2014},
  publisher={ACM New York, NY, USA}
}

@incollection{gates2000reconstruction,
  title={Reconstruction of cortical circuitry},
  author={Gates, Monte A and Fricker-Gates, Rosemary A and Macklis, Jeffrey D},
  booktitle={Progress in brain research},
  volume={127},
  pages={115--156},
  year={2000},
  publisher={Elsevier}
}

@article{arlotta2014brains,
  title={Brains in metamorphosis: reprogramming cell identity within the central nervous system},
  author={Arlotta, Paola and Berninger, Benedikt},
  journal={Current opinion in neurobiology},
  volume={27},
  pages={208--214},
  year={2014},
  publisher={Elsevier}
}

@article{castor2018resilience,
  title={Resilience after a neurological pathology: What impact on the cognitive abilities of patients with brain damage?},
  author={Castor, Naomie and El Massioui, Farid},
  journal={Neuropsychological rehabilitation},
  pages={1--19},
  year={2018},
  publisher={Taylor \& Francis}
}

@misc{firesmith_2019, title={System Resilience: What Exactly is it?}, url={https://insights.sei.cmu.edu/sei_blog/2019/11/system-resilience-what-exactly-is-it.html}, journal={System Resilience: What Exactly is it?}, author={Firesmith, Donald}, year={2019}, month={11}}

@article{lecun1998gradient,
  title={Gradient-based learning applied to document recognition},
  author={LeCun, Yann and Bottou, L{\'e}on and Bengio, Yoshua and Haffner, Patrick},
  journal={Proceedings of the IEEE},
  volume={86},
  number={11},
  pages={2278--2324},
  year={1998},
  publisher={Ieee}
}

@article{simonyan2014very,
  title={Very deep convolutional networks for large-scale image recognition},
  author={Simonyan, Karen and Zisserman, Andrew},
  journal={arXiv preprint arXiv:1409.1556},
  year={2014}
}
\end{document}